\documentclass{bmvc2k}

\title{PS-Transformer: Learning Sparse Photometric Stereo Network using Self-Attention Mechanism}

\addauthor{Satoshi Ikehata}{https://satoshi-ikehata.github.io/}{1}

\addinstitution{
 National Institute of Informatics\\
 Tokyo, JAPAN
}

\runninghead{Prof}{Satoshi Ikehata}

\def\ie{\emph{i.e}\bmvaOneDot}
\def\eg{\emph{e.g}\bmvaOneDot}

\def\etal{\emph{et al}\bmvaOneDot}

\newcommand{\Eref}[1]{Eq.~(\ref{#1})}
\newcommand{\Fref}[1]{Fig.~\ref{#1}}

\usepackage{bm}
\usepackage{times}
\usepackage{epsfig}
\usepackage{graphicx}
\usepackage{amsmath}
\usepackage{amssymb}
\usepackage{color}

\begin{document}

\maketitle

\begin{abstract}
Existing deep calibrated photometric stereo networks basically aggregate observations under different lights based on the pre-defined operations such as linear projection and max pooling. While they are effective with the dense capture, simple first-order operations often fail to capture the high-order interactions among observations under small number of different lights. To tackle this issue, this paper presents a deep sparse calibrated photometric stereo network named {\it PS-Transformer} which leverages the learnable self-attention mechanism to properly capture the complex inter-image interactions. PS-Transformer builds upon the dual-branch design to explore both pixel-wise and image-wise features and individual feature is trained with the intermediate surface normal supervision to maximize geometric feasibility. A new synthetic dataset named CyclesPS+ is also presented with the comprehensive analysis to successfully train the photometric stereo networks. Extensive results on the publicly available benchmark datasets demonstrate that the surface normal prediction accuracy of the proposed method significantly outperforms other state-of-the-art algorithms with the same number of input images and is even comparable to that of dense algorithms which input 10$\times$ larger number of images. 
\end{abstract}

\section{Introduction}
\label{sec:intro}
Photometric Stereo is a long-standing problem to recover a fine surface normal map from HDR (High Dynamic Range) images captured under different lights with a fixed camera. Since Woodham~\cite{Woodham1980} proposed the first Lambertian calibrated photometric stereo algorithm, optimization based inverse rendering had been a mainstream approach~\cite{Goldman2005, Alldrin2007a,Ikehata2012,Ikehata2014a,Shi2014}. However, recent breakthroughs from deep neural networks have sparked great interest to explore the data-driven photometric stereo algorithms to handle complex global illumination effects that cannot be described in a mathematically tractable form. 

Unlike most computer vision tasks, a photometric stereo network must accept a {\it set input}~\cite{Lee2019} (\ie, unordered, varying number of pairs of image and light) and the output should take into account the entire input set. Existing deep photometric stereo networks satisfied this requirement by mainly two permutation invariant aggregation methods, which are {\it observation-map}~\cite{Ikehata2018, Li2019, Zheng2019} and {\it set-pooling}~\cite{Chen2018,Chen2019,Chen2020}. The former introduces the fixed-shape 2-d map called {\it observation map} where all the observations at a single pixel are projected according to light directions to be fed to convolutional neural networks (CNN) for pixel-wise surface normal prediction. On the other hand, the latter is based on the set-pooling method~\cite{Zaheer2017} where a feature map is firstly encoded from a pair of image and  light, then all the feature maps from different images are merged using a pooling operation (\eg, mean, max) to be fed to CNN to predict the 2-d surface normal map. 

While both approaches accept a set input of varying size, recent follow-up works~\cite{Li2019,Zheng2019,Yao2020} which compared these two approaches pointed their drawbacks. Given enough input images, an observation map captures shading variations of individual pixels better  than set-pooling algorithms, however its performance significantly drops in the {\it sparse} photometric stereo setup  (\ie, the number of input images is small e.g., less than $10$) where shading variations of individual pixels are not sufficient to recover surface details. Since photometric stereo data acquisition requires large labor, it is more convenient to recover accurate normal map with the least number of images and the set-pooling algorithms have an advantage here owing to the feature map which encodes the intra-image shading information. To get the best of both approaches, Yao~\etal~\cite{Yao2020} have recently proposed a two-step approach named {\it GPS-Net}, which firstly aggregated the per-pixel shading variations using the trainable structure-aware graph convolution (SGC)~\cite{Chang2018}, then constructed a feature map by merging features of different pixels to be fed to CNN-based surface normal predictor to account for the intra-image spatial information. However, there are two problems in this method. First, their SGC filters constructed the neighborhood structure only from the virtual central node to each observation of the same pixel therefore incorporated only the first-order proximity. More recently, Liu~\etal~\cite{Liu2021} have proposed {\it SPS-Net} which firstly introduced the self-attention mechanism~\cite{Vaswani2017} to encode higher-order interactions among observations under different lights. The self-attention does not assume the fixed-size input nor regular grid structure therefore doesn't encounter the sparsity problem in the observation map. SPS-Net repeats a set of the self-attention block and some convolutions repeatedly with different image scales and finally performs the max-pooling to feed the features to the CNN-based normal map predictor. While the self-attention mechanism in SPS-Net may capture high-order interactions among observations, the intra-image spatial feature and inter-image photometric feature were intermediately jumbled up in the model via multiple convolutions and 
accompanied downsampling, therefore important perpixel shading information was not maximally utilized. In addition, SPS-Net aggregated all the information via the max-pooling at last, which could squash the higher order information before the final normal map prediction.

Based on these insights, this paper presents a transformer-based photometric stereo network, namely {\it PS-Transformer} that overcomes limitations in existing deep photometric stereo networks. The main ideas are two-fold. First, as with~\cite{Liu2021}, the self-attention mechanism in the transformer model~\cite{Vaswani2017} is introduced to aggregate features under different lights, however, multiple, multi-head self-attention layers are stacked in a row rather than alternating the self-attention and convolutions as in~\cite{Liu2021} to keep position specific information. Second, we introduce the dual-branch design to discriminate inter-image and intra-image information explicitly at the feature aggregation phase and fuse them just before the final prediction. In addition, the new synthetic training dataset, namely {\it CyclesPS+}, is also presented with training strategy optimized for it. It will be shown that PS-Transformer trained on CyclesPS+ dataset could dramatically improve the performance especially in the sparse photometric stereo problem, which is almost comparable to the performance on the dense input set.\\\\
\noindent \textbf{Preliminaries}: The goal of the calibrated photometric stereo is to recover a surface normal map ($N\in \mathbb{R}^{h\times w\times 3}$) from images ($I_1,\cdots,I_m\in \mathbb{R}^{h\times w\times c}$) captured under known directed lights ($\bm{l}_1,\cdots,\bm{l}_m|\bm{l}_j=[l_j^x,l_j^y,l_j^z]\in \mathbb{R}^{3})$ with a known fixed camera ($\bm{v}=[0,0,1]^\top$) where $w$, $h$ and $c$ are width, height and color channel of the image and $m$ is the total number of different lights. Henceforth $j$ indicates the light index and $i$ indicates the pixel index (\eg, $\bm{x}_{j,i}$ is the value at $i$-th pixel of $j$-th matrix $\bm{x}_j$). We assume that pixel intensities are normalized by their power of light, therefore $\bm{l}$ is supposed to be a unit vector.  
\section{Related Works} \label{sec:agg}
\subsection{Conventional Photometric Stereo Algorithms}
Before the advent of deep learning, most conventional photometric stereo algorithms recovered surface normals of a scene via a simple diffuse reflectance modeling (\eg, Lambertian) while treating other effects as outliers~\cite{Wu2010,Ikehata2012,Ikehata2014a,Queau2017}. While effective, a drawback of this approach is that if it were not for dense diffuse inliers, the estimation fails. To handle dense non-Lambertian reflections, various algorithms arrange the parametric or non-parametric models of non-Lambertian BRDF~\cite{Goldman2005,Alldrin2008,Shi2014, Ikehata2014b}. Even though nonlinear models can be applied to a variety of materials, they were powerless against model outliers. A few amount of photometric stereo algorithms are grouped into the example-based approach, which takes advantages of the surface reflectance of objects with known shape, captured under the same illumination environment with the target scene~\cite{Silver1980,Hui2017}. While effective, this approach also suffers from model outliers and has a drawback that the lighting configuration of the reference scene must be taken over at the target scene. 

\subsection{Deep Photometric Stereo Algorithms}
Deep photometric stereo networks basically consist of two modules (a) permutation invariant feature aggregation module and (b) surface normal prediction module. Given a set of input images and corresponding lights, the permutation invariant feature aggregation module encodes pixel-wise features as
\begin{equation}
\bm{f}_i = {\rm Agg}\{\bm{x}_{1,i}, \cdots, \bm{x}_{m,i}\}.\label{eq:agg}\\
\end{equation}
Here, $\bm{f}_i$ is a $d_{agg}$-dimensional vector ($d_{agg}$ varies by algorithm) and $\bm{x}_{j,i}$ is the feature from $i$-th pixel under $j$-th light and ``Agg'' is the aggregation over features at the same pixel under different lights in the permutation invariant manner. Both the aggregation operation and feature type differ between algorithms. In the observation-map based algorithm~\cite{Ikehata2018,Li2019,Zheng2019}, $\bm{x}_{j,i} \triangleq [I_{j,i},\bm{l}_j]$ and the aggregation operation is a set of projections of $I_{j,i}$ onto $l_x$-$l_y$ coordinates of the fixed size observation map. On the other hand, the set-pooling based algorithms~\cite{Chen2018,Chen2019,Chen2020} define $\bm{x}_{j,i} \triangleq \phi(I_j,\bm{l}_j)_i$ where $\phi(I_j,\bm{l}_j)$ is the feature map extracted from the image and light $\{I_j,\bm{l}_j\}$ and the aggregation is pixel-wise max or mean pooling of feature maps. GPS-Net~\cite{Yao2020} also defines $\bm{x}_{j,i} \triangleq [I_{j,i},\bm{l}_j]$ but the aggregation is based on the trainable SGC filters. Given aggregated features, the surface normal is predicted by either form of $N_i = \psi(\bm{f}_i)$ or $N = \Psi(\bm{f}_1,\cdots, \bm{f}_{wh})$. Here, $\psi$ is the pixel-wise surface normal predictor (\ie, a feed-forward neural network) and $\Psi$ is the image-wise surface normal predictor (\ie, a convolutional neural network) where individual features merged into a single feature map before the prediction. 
\begin{figure*}[!t]
	\begin{center}
		\includegraphics[width=120mm]{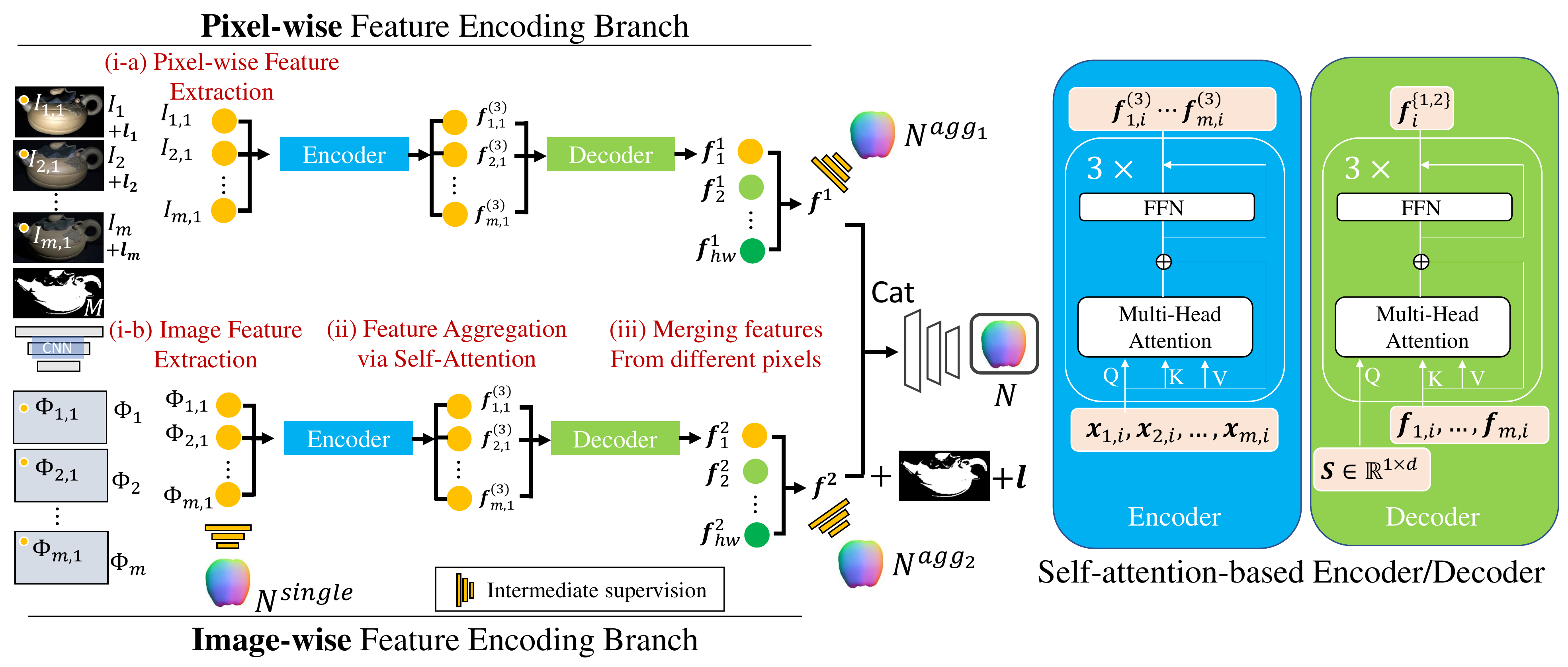}
	\end{center}
	\caption{The illustration of the PS-Transformer. Given $m$ number of input images with corresponding lights, the input is (i-a) directly passed to the pixel-wise feature encoding module for taking advantage of pixel-wise shading variations and also passed to the image-wise feature encoding module (i-b) after the feature maps are extracted from each image independently for taking advantage of spatial information. In each module, (ii) per-pixel feature vectors under different lights are aggregated by the encoder and the decoder based on the multi-head self-attention blocks and (iii) the aggregated feature vectors are merged into the single feature map by another self-attention layer to be fed to the CNN-based surface normal map predictor. To link intermediate features with surface normal information, the intermediate surface normal supervision is introduced when training the network.}
	\label{fig:architecture}
\end{figure*}
\section{PS-Transformer}
In this section, we introduce our transformer-based photometric stereo network, called {\it PS-Transformer}. The entire architecture is illustrated in~\Fref{fig:architecture}.
\subsection{Feature Aggregation using Self-Attention}
We first describe the feature aggregation operation (\ie, Agg in~\Eref{eq:agg}) in PS-Transformer which is the core component in our framework. Similar to other transformer-based architectures, our feature aggregation consists of an encoder network (\ie, concurrently encode the whole set) followed by a decoder network (\ie, pooling encoded features) as follow
\begin{equation}
\bm{f}_i = {\rm Decoder({\rm Encoder}\{\bm{x}_{1,i}, \cdots, \bm{x}_{m,i}\})}.\label{eq:ouragg}\\
\end{equation}
Before digging into the details, we first review the self-attention layer in the transformer model~\cite{Vaswani2017}. Given a set of features ($\bm{x}_{j}|1\leq j\leq m$), the self-attention layer firstly projects each item onto three different vectors: the query vector $W^Q\bm{x}_{j}$, the key vector $W^K\bm{x}_{j}$ and the value vector $W^V\bm{x}_{j}$ with embedding dimension $d_q, d_k, d_v$. Vectors computed from different items are then packed together into three different matrices, namely, $Q\in \mathbb{R}^{m\times d_q}$, $K\in \mathbb{R}^{m\times d_k}$ and $V\in \mathbb{R}^{m\times d_v}$. The output $A(Q,K,V)\in \mathbb{R}^{m\times d_v}$ of the self-attention layer is given by,
\begin{equation}
A (Q,K,V) = {\rm softmax}\left(\frac{QK^\top}{\sqrt{d_k}}V\right). \label{eq:selfattention}
\end{equation}
The self-attention layer computes the attention scores as normalized dot-product of the query with all keys and outputs the weighted sum of values where a value gets more weight if its corresponding key is similar to the query. Note that as derived from~\Eref{eq:selfattention}, the self-attention layer is invariant to permutations and changes in the number of input items. Since a single-head self-attention layer limits its ability to focus on simple interaction among items, it is common to comprises multiple self-attention blocks (\ie, multi-head attention). In this work, we also use a $h$-head multi-head attention layer ($h=8$ in our evaluation) and stack multiple multi-head attention layers in a row to encode higher order interactions. 

Using this self-attention layer, Encoder(.) in~\Eref{eq:ouragg} is represented by three sets of multi-head attention and feed-forward networks (FFN) as
\begin{eqnarray}
F_i^{(t)} &\triangleq& [\bm{f}^{(t)}_{0,i}\cdots \bm{f}_{m,i}^{(t)}]^{\top} \in \mathbb{R}^{m\times d}\nonumber\\
Q &=& F_i^{(t)}W^Q \in \mathbb{R}^{m\times d_q},\;K = F^{(t)}_iW^K\in \mathbb{R}^{m\times d_k},\;V = F^{(t)}_iW^V\in \mathbb{R}^{m\times d_v}, \nonumber\\ 
H &=& Q + {\rm MultiheadAttn}(Q, K, V), \nonumber\\
F_i^{(t+1)} &=& {\rm FFN}({\rm GeLU}({\rm FFN}(H))) + H,
\end{eqnarray}
where $\bm{f}^{(t)}_{j,i}\in \mathbb{R}^d$ is the feature vector at $i$-th pixel under $j$-th light (\ie, $\bm{f}^{(0)}_{j,i} = \bm{x}_{j,i}$) and $W^Q$, $W^K$ and $W^V$ are the projection matrices and GeLU is the Gaussian error Linear Units~\cite{Hendrycks2016}. Note that we use the dropout (the probablity is $0.1$) after the activation but does not use the Layer Normalization (LayerNorm) which is commonly used in the exisiting transformer models because we empirically found it degrades the performance. The dimensionality of the hidden layers is fixed by $d=d_q=d_k=d_v=256$ throughout the manuscript.

Given the input features $\{\bm{x}_{1,i}, \cdots, \bm{x}_{m,i}\}$, the output of the encoder is a set of encoded features contained in the matrix $F_i^{(3)} = [\bm{f}^{(3)}_{1,i} \cdots \bm{f}^{(3)}_{m,i}]^{\top}\in \mathbb{R}^{m\times d}$. For aggregating the information over different lights, we then apply the Decoder(.) in~\Eref{eq:ouragg} to shrink the feature size without losing the interactions among set items as is independent of the input set size $m$. To achieve this, we introduce the PMA (Pooling by Multihead Attention) module~\cite{Lee2019} which applies multi-head attention on a learnable seed vector $S\in \mathbb{R}^{1\times d}$ as key vector and use $F_i^{(3)}$ as query and value vectors to get perpixel $d$-dimensional vectors $\bm{f}_i$. The next section details our photometric stereo network architecture that uses this feature aggregation operation. 
\subsection{Network Architecture}
As have already been discussed in previous works~\cite{Li2019,Zheng2019,Yao2020}, the intensity variations of individual pixels over different lights contribute to recover the sharp geometric boundaries and local spatial intensity variations among neighbor pixels contribute to recover geometry from a sparse input set. To get the best of them, PS-Transformer introduces the dual-branch design as depicted in~\Fref{fig:architecture}. Both branches consist of self-attention based feature aggregation as detailed and the only difference in two branches is the type of input features to be aggregated. Concretely, the first branch takes $\bm{x}^1_{j,i} \triangleq [I_{j,i},\bm{l}_j]\in \mathbb{R}^{c+3}$ and the second branch takes $\bm{x}^2_{j,i} \triangleq [\phi(I_j,M)_i,\bm{l}_j]\in \mathbb{R}^{67}$ as input where $M\in \mathbb{R}^{h\times w}$ is the object mask which gives the network the boundary constraint as is known to be helpful in the shape-from-shading literature~\cite{Zhang1999}\footnote{Object mask is simply acquired by taking the non-zero (or more than some small value for the robustness) pixels in the image averaged over all the light directions.}. $\phi$ is shared-weight convolutional neural networks with six $3\times 3$ convolution layers, normalization (Batch Normalization) and activation (Leaky-ReLU) to output the sixty four dimensional feature map. The acquired feature map is then concatenated with a lighting map where $\bm{l}_j\in\mathbb{R}^{3}$ is expanded to the size of the image. Note that the encoder/decoder in two branches don't share the network parameters. The output from these two branches are concatenated with the object mask, then a feature map of $\mathbb{R}^{h\times w\times (2d+1)}$ is fed to image-space surface normal predictor $\Psi$ which consists of five $3\times 3$ convolution layers whose number of filters are $2d+1$ except for the final surface normal prediction layer to get $N\in \mathbb{R}^{h\times w \times 3}$. In summary, our PS-Transformer architecture is formally described as
\begin{eqnarray}
\bm{f}^1_i &=& {\rm Decoder_1}({\rm Encoder_1}\{\bm{x}^1_{1,i}, \cdots, \bm{x}^1_{m,i}\}),\;\;\;\bm{x}^1_{j,i} \triangleq [I_{j,i},\bm{l}_j],\nonumber\\
\bm{f}^2_i &=& {\rm Decoder_2}({\rm Encoder_2}\{\bm{x}^2_{1,i}, \cdots, \bm{x}^2_{m,i}\}),\;\;\;\bm{x}^2_{j,i} \triangleq [\phi(I_j,M)_i,\bm{l}_j],\label{eq:ourmodel}\\
N &=& \Psi({\rm cat}\{\bm{f}^1_1,\bm{f}^2_1,M_1\},\cdots,{\rm cat}\{\bm{f}^1_{hw},\bm{f}^2_{hw},M_{hw}\}).\nonumber
\end{eqnarray}
where ``cat'' is the operation for the pixel-wise concatenation.

\subsection{Loss Function}
The network is trained with a simple mean squared loss between predicted and ground truth surface normal maps. In addition to evaluating the final output ($N$), we also put intermediate surface normal supervision to encourage each feature to directly associate with the surface normal prediction. Concrete form of the loss function is as follow,
\begin{eqnarray}
L_{PST} &=& \|M\odot (N-N^{gt})\|_2 + \frac{1}{m}\sum_{j=1}^m\|M\odot (N_{j}^{single}-N^{gt})\|_2 \\
&+& \|M\odot (N^{{agg}_1}-N^{gt})\|_2 +  \|M\odot (N^{{agg}_2}-N^{gt})\|_2,
\end{eqnarray}
where $\odot$ denotes Hadamard product to remove background pixels from the loss computation. Here, $N_j^{single}$ is the surface normal map predicted from the $j$-th single-view feature map $\bm{x}^2_j\in \mathbb{R}^{h\times w\times 67}$ in~\Eref{eq:ourmodel} by six $3\times 3$ convolution layers whose number of filters is $67$ except for the final output layer with normalization (Batch Normalization) and activation (Leaky-ReLU). $N^{{agg}_{\{1,2\}}}$ are surface normal maps formed by predicted surface normal vectors from aggregated features ($\bm{f}^1_i$ and $\bm{f}^2_i$ $|1\leq i \leq hw$) by two fully-connected layers (the dimension changes as $d$→$d$→3) with activation (Learky-ReLU) to predict three dimensional normal vector. We assigned equal weight to the contribution of each term without valuing specific normal map strongly. 
\section{Results}
\noindent \textbf{Algorithms}: PS-Transformer is evaluated with representative photometric stereo networks based on the observation map (CNN-PS~\cite{Ikehata2018}), set-pooling (PS-FCN+~\cite{Chen2020})\footnote{PS-FCN+~\cite{Chen2020} is the extension from PS-FCN~\cite{Chen2018} where data normalization strategy to equalize spatial appearance has been introduced.} and the graph convolution (GPS-Net~\cite{Yao2020}). We also compared our method against {\it SPS-Net}~\cite{Liu2021} where the self-attention mechanism is also applied to interact features under different lights. 

In our experiments, we used authors' official implementations and pretrained models~\cite{CNNPS,PSFCN,GPSNET} with minor modifications to evaluate them with the same training/test protocol~\footnote{Please see supplementary materials for further details.}. For the fair comparison, we also compared against existing models trained on our training data which results in five competitors in total; (i) PS-FCN+ (Trained on our dataset), (ii) PS-FCN+ (Pretrained), (iii) GPS-Net (Trained on our dataset), (iv) GPS-Net (Pretrained), (v) CNN-PS (Trained on our dataset)\footnote{We didn't use the pretrained model of CNN-PS since the public CNN-PS pretrained model was trained on large number of input images (\ie, $m\geq 30$). As has already been mentioned in~\cite{Chen2020}, when the number of images between training and test is largely different, the performance significantly drops.} and (iv) SPS-Net (Trained on our dataset). The original implementation of SPS-Net was trained on $32\times 32$ patches of the Blobby and Sculpture datasets~\cite{Chen2018}, however it is not clear if this patch size is optimal for our CyclesPS+ dataset which will be described later. Therefore we compared two different configurations of SPS-Net where the network was trained on either of $32\times 32$ or $8\times 8$ (same training patch size as PS-Transformer) patches for the fair comparison. \\\\
\noindent \textbf{Training Dataset}: The existing deep photometric stereo models were trained either on Blobby shape datasets~\cite{Chen2018} or CyclesPS datasets~\cite{Ikehata2018}. The Blobby shape dataset contains much bigger number of samples ($\ie, 85212$), however the material is spatially uniform and lighting variations in each dataset is small (\ie, 64). On the other hand, CyclesPS datasets only contain 15 different objects but the material is spatially varying and the lighting variations in each dataset is large (\ie, 740). To get the best of them, we created {\it CyclesPS+} dataset following the rendering scheme in CyclesPS but increased the dataset size (25 objects including 15 objects in CyclesPS and different types of subsets for each object). Please refer supplementary materials for further instruction.
\\\\
\noindent \textbf{Training Details}:
Since we mainly target the sparse photometric stereo problem (\ie, $m\leq 10$), the number of lights in a training sample is always fixed by $10$ for training all the models including ours and competitors (except when using pretrained models). We should note that it has often been reported that the test accuracy improves when the number of images at the training phase matches with one for test~\cite{Chen2020}. However, it is quite inefficient to have models for every different number of images, therefore we reuse a single trained model (\ie, $m=10$) for a varying size of input set at test. Due to the space limit, the detailed training protocol (\eg, number of epochs or learning strategy) is presented in the supplementary. 
\\\\
\textbf{Test Dataset}: Our main result is based on DiLiGenT~\cite{Shi2018} and DiLiGenT-MV~\cite{Shi2020}. The number of real objects in total is $110$ ($10$ objects  $\times$ $1$ view~\cite{Shi2018} and $5$ objects $\times$ $20$ views~\cite{Shi2020}) which is significantly larger than all the existing evaluations in~\cite{Ikehata2018,Chen2020,Yao2020} where only DiLiGenT~\cite{Shi2018} was used. Each data provides 16-bit integer HDR images with a resolution of 612×512 from 96 different known lighting directions. The ground truth surface normals for the orthographic projection and the single-view setup are also provided. The inference of models for sparse photometric stereo is basically unstable according to the light distribution (\ie, the condition number is large), therefore we performed 10 random trials and averaged them. Specifically, ten sets of $m$ random light directions were sampled from the upper hemispherical surface in advance, and {\it the same sets of the light distribution were used for all methods}. .\footnote{The effect of the different light distribution is discussed in the supplementary material.} We also note that we didn't include the results on synthetic data in our work, however important analysis on synthetic data such as the study about the ability of Principled BRDF~\cite{DisneyPrincipledBSDF} in representing the real materials have already been provided in the previous work~\cite{Ikehata2018}. 
\\\\ 
\begin{figure}[!t]
    \begin{tabular}{cc}
      \begin{minipage}[t]{0.46\linewidth}
        \centering
        \includegraphics[keepaspectratio, scale=0.25]{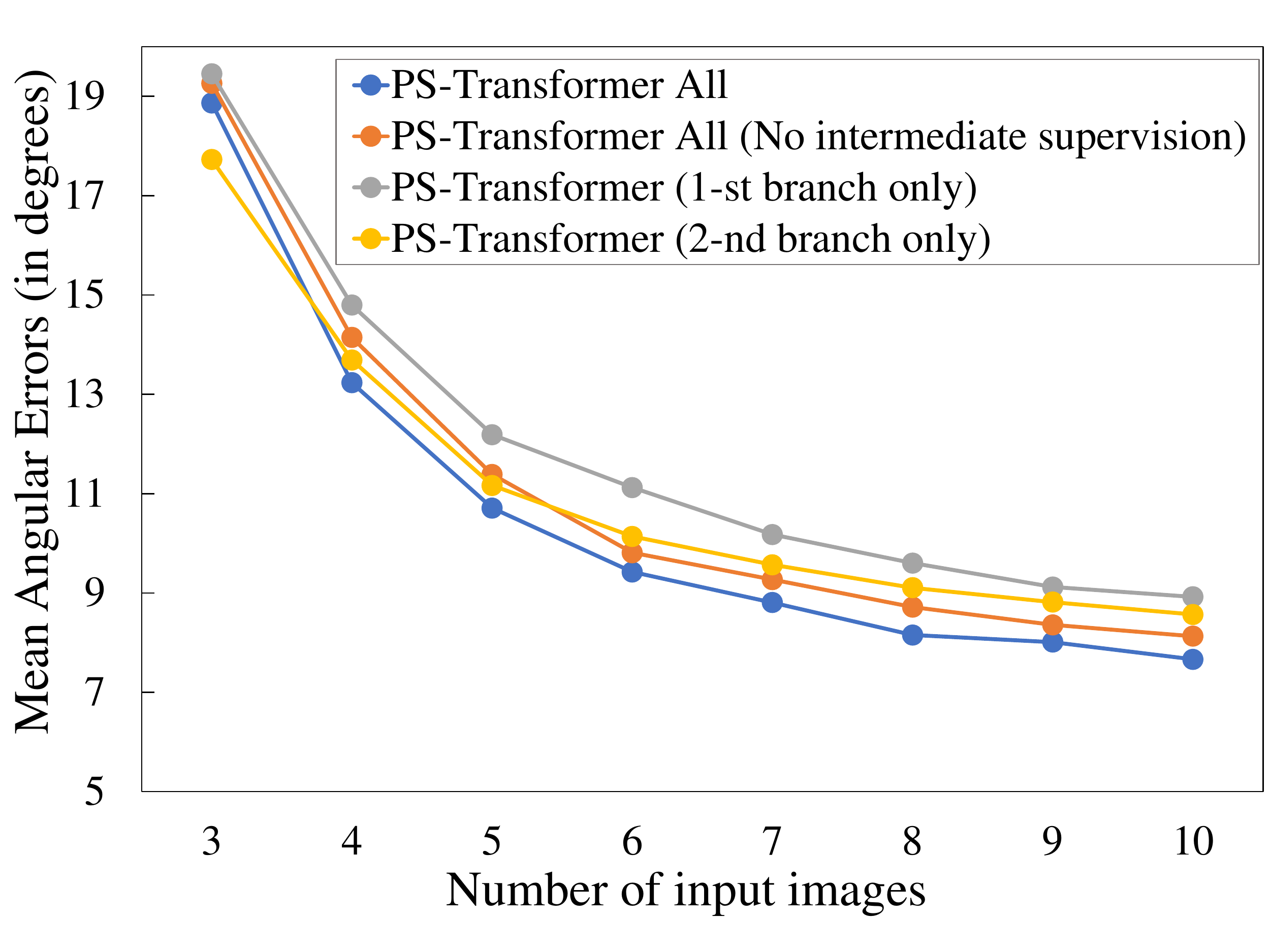}
        \caption{Ablation analysis to justify the design of our architecture. Errors for 10 objects are averaged.}
        \label{fig:ablation}
      \end{minipage} &
      \begin{minipage}[t]{0.45\linewidth}
        \centering
        \includegraphics[keepaspectratio, scale=0.25]{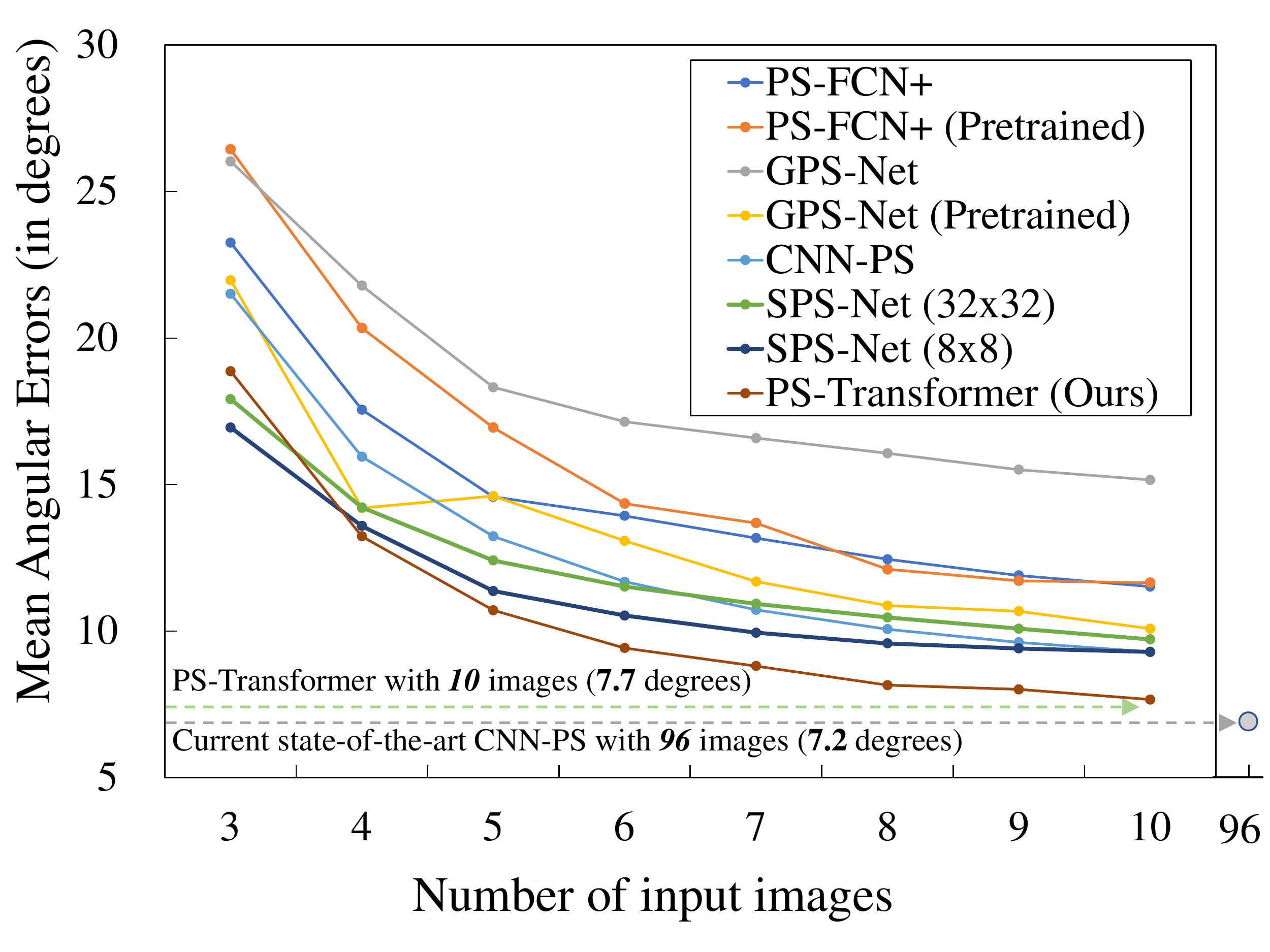}
        \caption{The quantitative comparison on DiLiGenT dataset. Errors for 10 objects are averaged.}
        \label{fig:teaser}
      \end{minipage}
    \end{tabular}
  \end{figure}

\begin{figure}[!t]
    \begin{tabular}{cc}
      \begin{minipage}[t]{0.45\linewidth}
        \centering
        \includegraphics[keepaspectratio, scale=0.25]{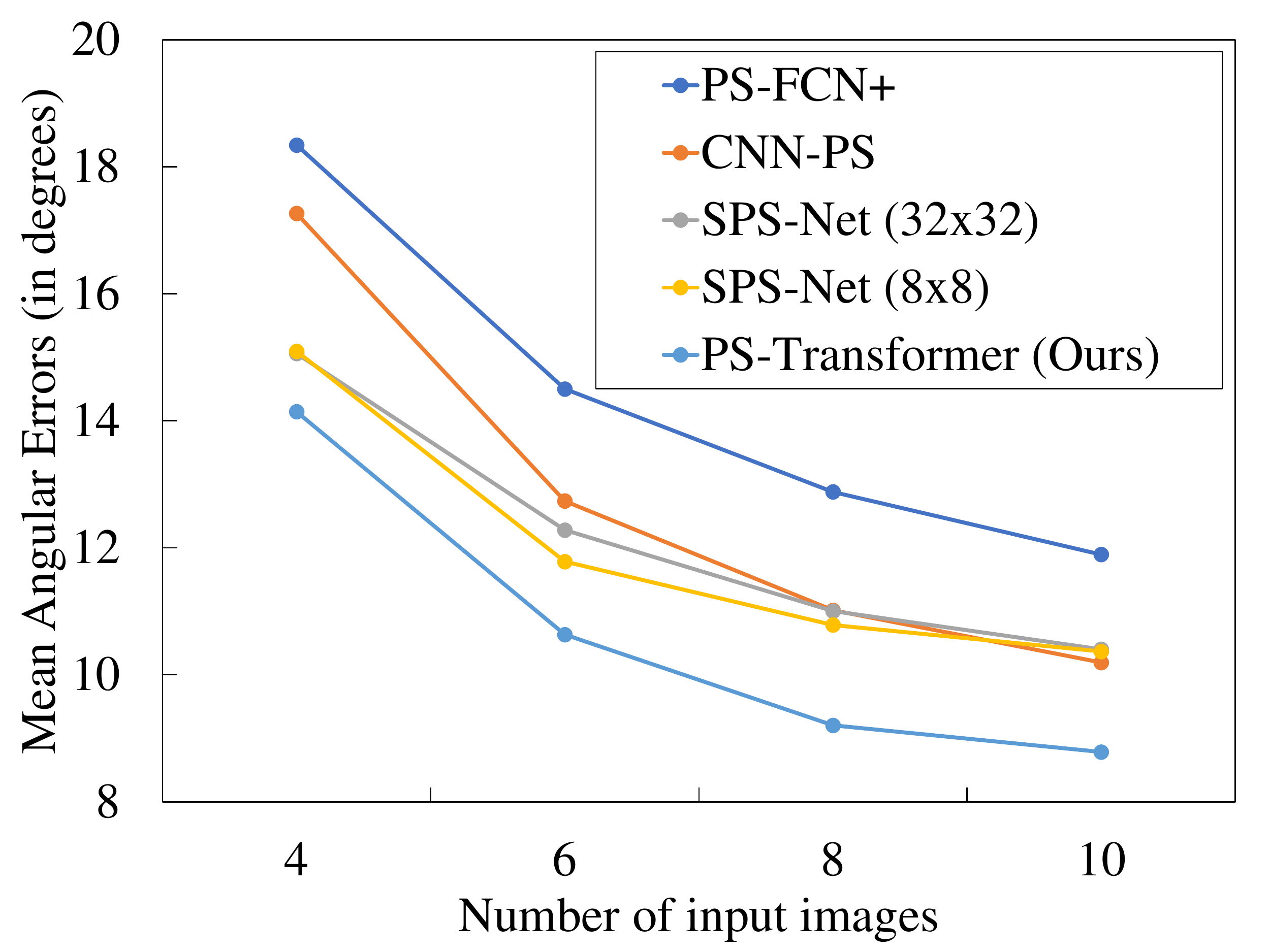}
        \caption{The quantitative comparison on DiLiGenT-MV dataset. Errors for five objects and twenty views are averaged.}
        \label{fig:diligentmv}
      \end{minipage} &
      \begin{minipage}[t]{0.45\linewidth}
        \centering
        \includegraphics[keepaspectratio, scale=0.25]{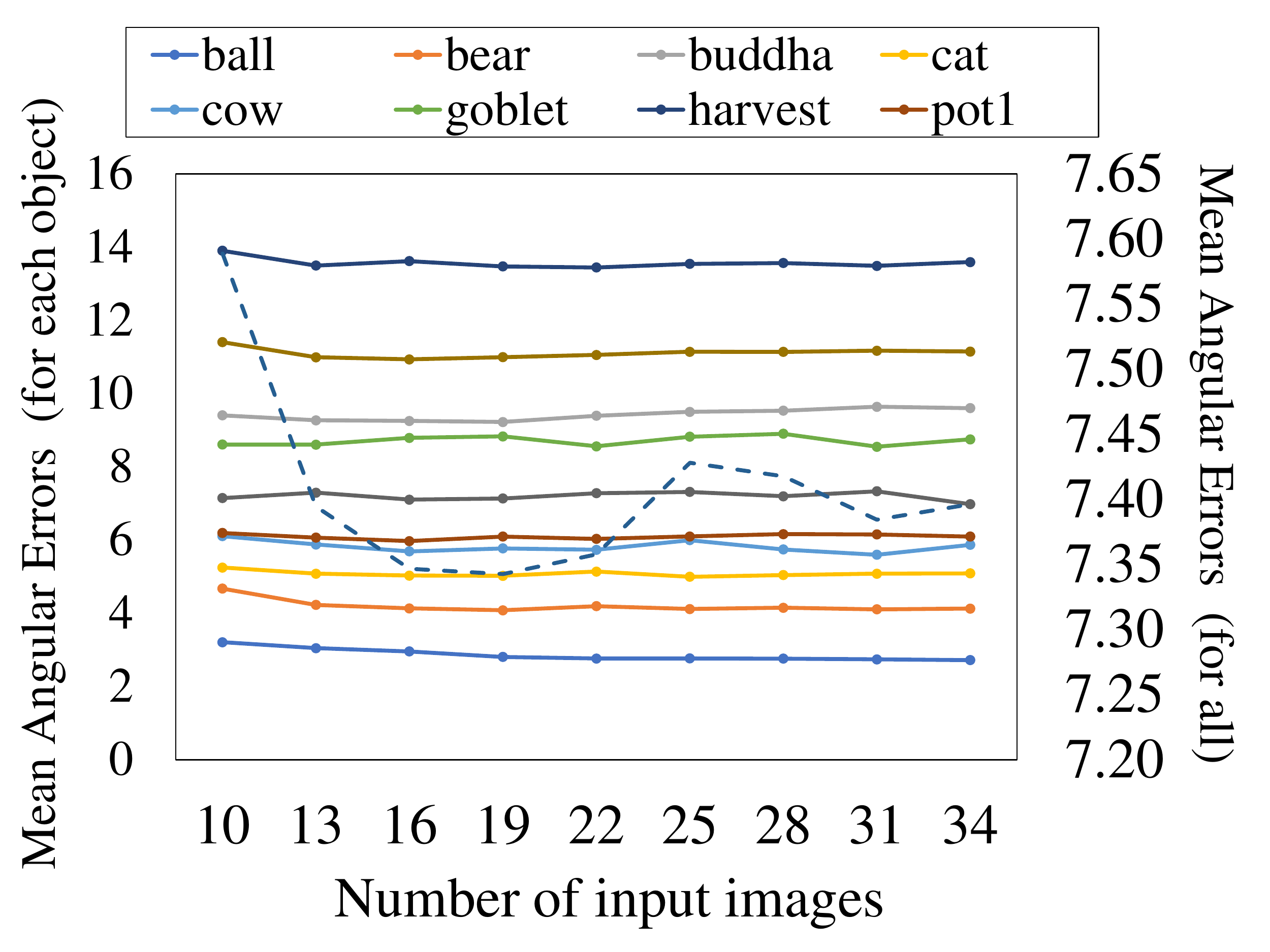}
        \caption{The sutdy with test images whose number is more than 10.}
        \label{fig:morethan_10}
      \end{minipage}
    \end{tabular}
  \end{figure}
  
 \begin{figure}[!t]
	\begin{center}
		\includegraphics[width=120mm]{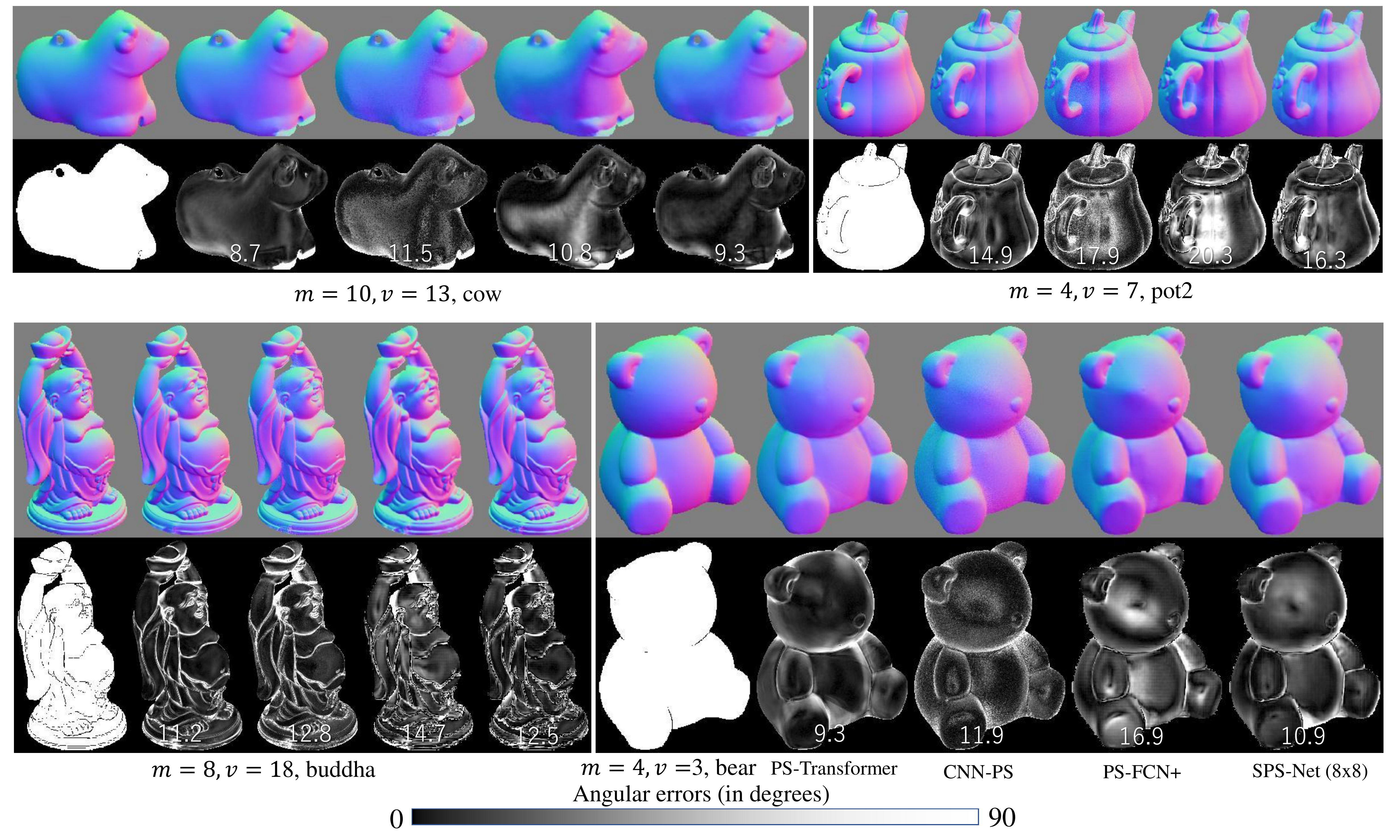}
	\end{center}
	\caption{Qualitative evaluation on DiLiGenT-MV dataset.}
	\label{fig:samples}
\end{figure}
\noindent \textbf{Ablation Study}:
We give ablation analysis to justify the design choice of PS-Transformer model including dual-branch design and the intermediate surface normal supervision. Here, we compared four different architectures on DiLiGenT dataset; (a) the architecture with a complete set of features (All) and (b) the same architecture but without the intermediate supervision, (c) the architecture only with the first branch (remove $\bm{f}^2$ from surface normal prediction module but with intermediate supervision), and (d) the architecture only with the second branch (remove $\bm{f}^1$ from surface normal prediction module but with intermediate supervision). The result is shown in~\Fref{fig:ablation}. Here, we showed the mean angular errors averaged over 10 DiLiGenT objects. We observed that PS-Transformer model with the dual-branch architecture showed the best performance as expected. It was also observed that the intermediate supervision improved the prediction accuracy. 
\\\\
\noindent \textbf{Quantitative Evaluation on DiLiGenT Dataset}:
We compared our method (full configuration) against five competitors mentioned in ''Algorithms'' on DiLiGenT main dataset in the sparse photometric stereo setup ($3\leq m \leq 10$). The results are illustrated in~\Fref{fig:teaser}. Here, the mean angular errors (MAE) averaged over 10 objects are presented\footnote{Please refer supplementary material to see prediction errors of individual objects.}. First, we observe that our method consistently outperforms other competitors in most cases. When the number of images is $10$, the accuracy is comparable to the current state-of-the-art {\it dense} photometric stereo algorithm (\ie, MAE = 7.2 with 96 images as reported in~\cite{Ikehata2018}) which shows the effectiveness of our transformer-based aggregation and dual-branch design. An interesting observation is that purely perpixel CNN-PS achieved roughly second best performance, which is even better than GPS-Net. This is contrary to the results in recent works~\cite{Li2019, Zheng2019,Yao2020} which claimed that a na\"ive observation map does not work well for the sparse photometric stereo setup. 

SPS-Net~\cite{Liu2021} also introduces the self-attention mechanism for interacting information under different lights, however this method has inferior performance to our method. One of the obvious reasons could be that SPS-Net inherit the image-wise feature maps extracted at very early stage to the end therefore less took advantage of the per-pixel shading variations. In addition, SPS-Net doesn't introduce the intermediate supervision where our ablation study has already proved that it significantly contributes to improve the performance. There are also many differences in the design of the architecture. SPS-Net repeats the {\it single} self-attention block (called PF-Block in~\cite{Liu2021}) and the convolution repeatedly with different image scales while our method stacks the multiple, multi-head self-attention in a row {\it without} changing the image scale which contributes to keep the shape surface details. Furthermore, SPS-Net simply performs the max-pooling operation after the PF-Block for aggregating the feature maps under different lights unlike ours using PMA module~\cite{Lee2019} for decoding feature maps in more data-adaptive manner. It is interesting to observe that the performance of SPS-Net ($8\times 8$) consistently outperforms SPS-Net ($32\times 32$) because it indicates that the photometric stereo networks work better when observing the local shading variations rather than observing more global information. This also supports the advantages of our dual-branch design that considers both local and global information.
\\\\
\noindent \textbf{Quantitative Evaluation on DiLiGenT-MV Dataset}:
The major drawbacks in DiLiGenT dataset is that the number of objects is only 10 and the variation of the surface normal distribution was quite limited. Recently, DiLiGenT dataset was extended to the multi-view edition as DiLiGenT-MV~\cite{Shi2020}. This new dataset contains images of 5 objects of complex BRDFs taken from 20 views (100 data in total). Here, we evaluated our method on DiLiGenT-MV with CNN-PS~\cite{Ikehata2018} and PS-FCN+~\cite{Chen2020} and SPS-Net ($32\times 32$ or $8\times 8$)~\cite{Liu2021} which were trained on our training dataset for the fair comparison~\footnote{We didn't include GPS-Net in this experiment because the authors implementation was optimized to DiLiGenT dataset and wasn't available on DiLiGenT-MV.}.The result is illustrated in~\Fref{fig:diligentmv} and~\Fref{fig:samples}. Because the number of data is 100 ($5$ objects $\times$ $20$ views), we averaged MAE over 100 data for the convenience\footnote{The prediction errors of individual objects and views as well as full qualitative comparison are provided in the supplementary material.}. The result is basically consistent with what we observed in the evaluation on DiLiGenT main dataset. When we observe the qualitative comparison in~\Fref{fig:samples}, our method consistently produced less noisy surface normal maps while preserving the surface details. We want to emphasize that all the algorithms including ours were trained on the same data and same strategy, therefore the difference simply comes from the architecture. This illustrates that how the attention mechanism from the transformer models and our two-branch model contributed to improve the performance in the sparse photometric stereo problem.
\\\\
\noindent \textbf{Performance on Dense Problem}: Though dense photometric stereo problem (\ie, $m\geq 10$) is out-of-scope of this work, we show the result of applying our model trained on $M=10$ to the test images whose number is larger than $10$. The result is illustrated in~\Fref{fig:morethan_10}.  We illustrate the mean angular error (in degrees) for both individual objects (\ie, solid line, left scale) and the average of 10 DiLiGenT objects (\ie, dash-line, right scale). In summary, we didn't observe the significant drop of the prediction accuracy when we input test images whose number is very different from one of training images. However, we didn't observe the significant improvement as well though much larger information is available. In reality, this result is not surprising and coincides with the observation in existing works~\cite{Chen2020,Yao2020} that models trained on the sparse input set are hard to be generalized to the dense setup. As described in the main paper, we need to increase the number of {\it training} images to represent more complex interactions among large number of input images. However, the transformer model is well known to be inefficient especially when the number of elements is large, so the adaptation of our model to the dense problem should be left for the future work. 

\section{Conclusion}
In this paper, we presented the self-attention-based photometric stereo network namely PS-Transformer. By incorporating the attention mechanism to capture the complex high-order interactions into the dual-branch designed network to capture both local and global information, our model significantly outperformed any existing deep photometric stereo algorithms in the sparse photometric stereo problem. The current limitation is that the self-attention requires O($M^2$) computation therefore our method doesn't scale to the very dense photometric stereo problem. However, developing efficient transformer models is the very hot topic and we believe our PS-Transformer models can benefit from it.
{\small
\bibliographystyle{ieee_fullname}
\bibliography{egbib}
}
\end{document}